\documentclass[runningheads]{llncs}
\usepackage[T1]{fontenc}
\usepackage{graphicx}
\usepackage{booktabs}
\usepackage[misc]{ifsym}
\newcommand{\corr}{(\Letter)}
\usepackage{paper}

\begin{document}

\title{Pointer-Guided Pre-Training: Infusing Large Language Models with Paragraph-Level Contextual Awareness}

\titlerunning{Pointer-Guided Pre-Training: Infusing LLMs with Paragraph Awareness}


\author{Lars Hillebrand\inst{1,2} \corr \and 
Prabhupad Pradhan\inst{1} \and
Christian Bauckhage\inst{1,2} \and \\
Rafet Sifa\inst{1,2}}
\authorrunning{L. Hillebrand et al.}
%
\institute{Fraunhofer IAIS, Germany \and
University of Bonn, Germany
\\
\email{lars.patrick.hillebrand@iais.fraunhofer.de}
}

\tocauthor{Lars Hillebrand, Prabhupad Pradhan, Christian Bauckhage, Rafet Sifa}
\toctitle{Pointer-Guided Pre-Training: Infusing Large Language Models with Paragraph-Level Contextual Awareness}

\maketitle              

\begin{abstract}
We introduce ``pointer-guided segment ordering'' (SO), a novel pre-training technique aimed at enhancing the contextual understanding of paragraph-level text representations in large language models. Our methodology leverages a self-attention-driven pointer network to restore the original sequence of shuffled text segments, addressing the challenge of capturing the structural coherence and contextual dependencies within documents. This pre-training approach is complemented by a fine-tuning methodology that incorporates dynamic sampling, augmenting the diversity of training instances and improving sample efficiency for various downstream applications.
We evaluate our method on a diverse set of datasets, demonstrating its efficacy in tasks requiring sequential text classification across scientific literature and financial reporting domains. Our experiments show that pointer-guided pre-training significantly enhances the model's ability to understand complex document structures, leading to state-of-the-art performance in downstream classification tasks. 
\end{abstract}
\section{Introduction}

The landscape of natural language processing (NLP) has been profoundly transformed by the emergence of generative large language models (LLM) such as OpenAI's GPT series \cite{achiam2023gpt,brown2020language}, Mixtral \cite{jiang2024mixtral}, and Llama2 \cite{touvron2023llama}. These models have set new benchmarks across a wide range of NLP tasks, showcasing remarkable capabilities in understanding and generating human language. Despite the significant advancements achieved by these large-scale models, there remains an equally important domain for smaller, specialized language models that excel in fast retrieval and semantic search, particularly those that generate precise paragraph and section representations. This domain is crucial, especially in the context of retrieval augmented generation (RAG) \cite{lewis2020retrieval}, where the integration of retrieval mechanisms with generative models enhances the reliability and informativeness of the output.

In this work, we address the important area of representation learning for improved paragraph-level contextual understanding, which is critical for enhancing the capabilities of NLP systems in sequential text classification and retrieval-based applications such as semantic text search.

\begin{figure}[t]
  \centering
  \includegraphics[width=\linewidth]{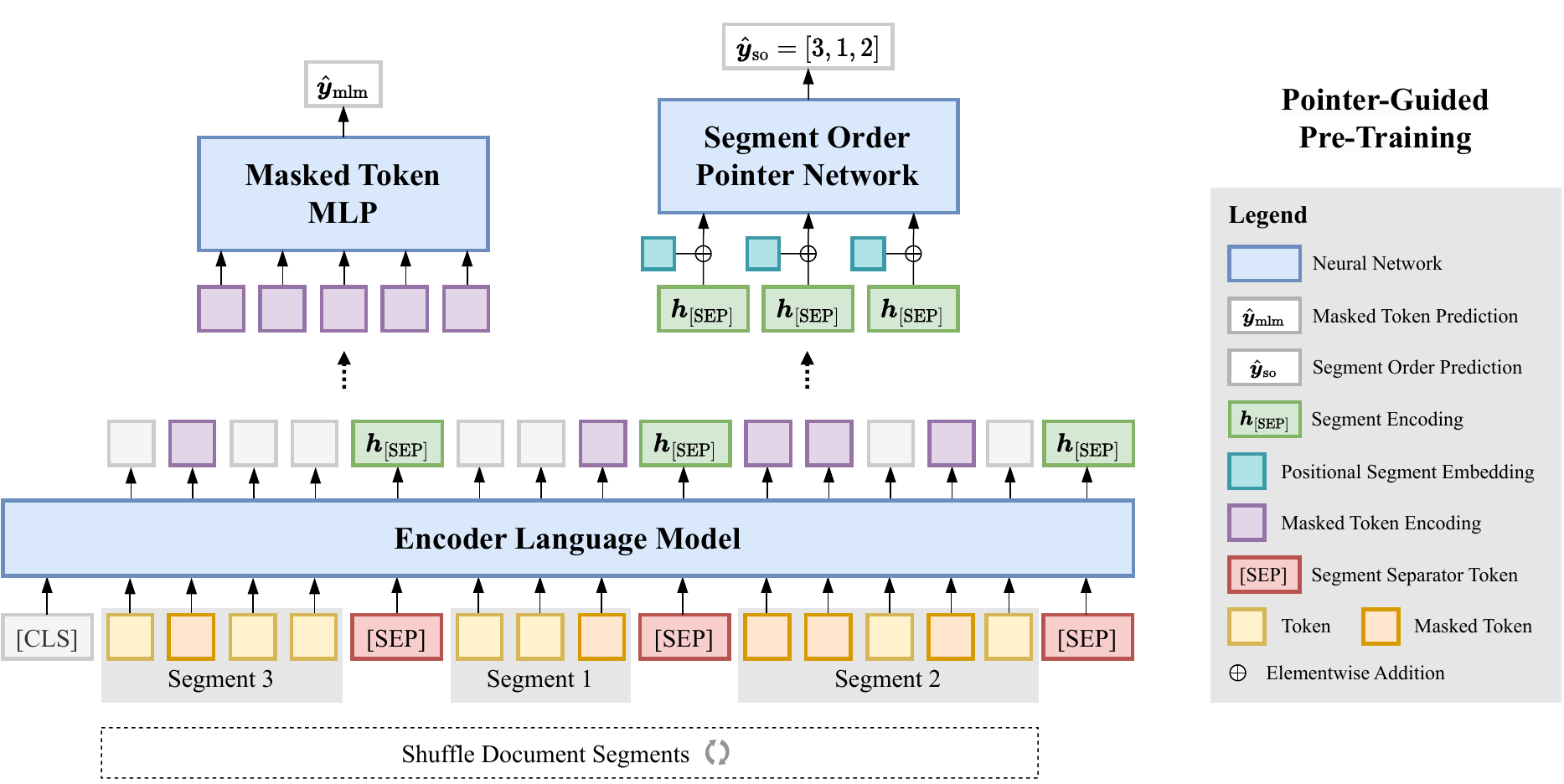}  
  \caption{Schematic visualization of our ``Pointer-Guided Pre-Training'' methodology. During pre-training a self-attention-based pointer network classification head learns to reconstruct the original order of shuffled text segments based on their hidden state representations ($\bm{h}_{\text{[SEP]}}$). Employing this segment ordering (SO) pre-training mechanism alongside masked language modeling (MLM) increases the segment level contextual awareness of the encoding language model and subsequently improves its downstream classification capabilities.}
  \label{fig:architecture}
\end{figure}

At the heart of our approach is the introduction of a novel pre-training methodology, named ``pointer-guided segment ordering'' (SO). This technique is designed to infuse language models with a deep awareness of paragraph-level context. Utilizing a self-attention-driven pointer network, the pointer-guided SO task challenges the model to reconstruct the original sequence of shuffled text segments (see Figure \ref{fig:architecture}). This complex task enables the model to develop a nuanced comprehension of narrative flow, coherence, and contextual relationships, significantly enhancing its ability to understand and represent paragraph-level context when combined with standard pre-training techniques like masked language modeling.

To complement our pre-training methodology, we further introduce dynamic sampling during the fine-tuning phase. This sampling technique increases the diversity of training instances across epochs, thereby improving sample efficiency. Dynamic sampling is particularly advantageous for smaller fine-tuning datasets characterized by long documents, where it effectively mitigates the risk of overfitting and promotes better generalization.

We demonstrate the effectiveness of these contributions through extensive experiments. We show that models pre-trained with our pointer-guided SO task consistently outperform competing baselines and raise the state-of-the-art across various datasets and tasks in the scientific literature and financial reporting domain. Furthermore, the model-agnostic nature of our methodology positions it to capitalize on future advancements in language model design, promising further improvements in paragraph-level text representation.

In summary, our work not only introduces a novel and effective methodology for enhancing paragraph-level embeddings but also establishes a new benchmark for sequential text classification. By opening new avenues for research in leveraging document structure for enhanced language modeling, our work marks a significant step forward in the ongoing evolution of NLP, with substantial potential impact on retrieval-based applications like semantic text search.

In the following, we first review related work, before describing our modeling approach in Section~\ref{sec:methodology}. In Section~\ref{sec:experiments}, we outline our experiments, describe our datasets, and discuss the results. Section~\ref{sec:conclusion} then draws a conclusion and provides an outlook into conceivable future work.

\section{Related Work}
\label{sec:related}

Traditional language modeling tasks such as masked language modeling (MLM) and next token prediction have been instrumental in learning token-level representations. Models like RoBERTa \cite{Liu2019RoBERTa}, ELECTRA \cite{Clark2020ELECTRA}, and GPT variants \cite{achiam2023gpt,brown2020language,touvron2023llama} have shown significant success in these areas. However, these models often lack mechanisms to enforce the learning of meaningful segment-level representations, crucial for understanding paragraph-level context. BERT \cite{Devlin19} introduced the next sentence prediction (NSP) task to bridge this gap, but its simplicity limited its effectiveness. Subsequent models, such as RoBERTa, abandoned NSP due to its limited contribution to model performance. Unlike these approaches, our work introduces a pointer-guided segment ordering methodology that directly leverages the inherent structure of textual data, offering a novel way to enhance paragraph-level understanding without relying on external data sources like Wikipedia article links in LinkBERT \cite{yasunaga2022linkbert} or knowledge-graph reasoning \cite{yasunaga2022deep}. The underlying architecture of our method, the pointer network \cite{NIPS2015_29921001}, has been successfully used for stand-alone sequence ordering tasks, as demonstrated by \cite{chowdhury2021everything,cui-etal-2018-deep,logeswaran2018sentence}. 
However, to the best of our knowledge we are the first to employ a novel self-attention-driven pointer network for segment ordering in conjunction with large language model pre-training. 

We evaluate our pre-training technique on several sequential text classification tasks. Previous studies have tackled these challenges with domain-adapted and fine-tuned BERT models \cite{cohan-etal-2019-pretrained,hillebrand_sustain} and the incorporation of hierarchical LSTMs, attention mechanisms, and CRF layers for improved sequential label dependency handling \cite{brack2022crossdomain,jin2018hierarchical,shang2021span,yamada2020sequential}.

\section{Methodology} \label{sec:methodology}

In this section, we provide a comprehensive description of our methodological contributions aimed at enhancing the contextual sensitivity of paragraph-level text representations, as well as their optimization for various downstream applications. Initially, we describe our novel ``pointer-guided segment ordering'' approach, a versatile pre-training strategy that employs a self-attention-driven pointer network to accurately restore the original sequence of shuffled text segments. Subsequently, we detail our fine-tuning methodology, which incorporates dynamic sampling to augment the diversity of training instances throughout successive training epochs, thereby improving sample efficiency.

\subsection{Pointer-guided Segment Ordering} \label{sec:segment-ordering}

A text document is inherently composed of consecutive text segments, which can range from whole paragraphs and individual sentences to enumerations, tables, and headlines. These segments are typically contextually interdependent, forming a coherent narrative in various types of documents, such as news articles, fiction novels, annual reports, or legal contracts.

To capture the essence of this structural coherence, we propose a novel self-supervised pre-training technique denoted as pointer-guided segment ordering (SO) that is capable of leveraging large amounts of unlabeled text data to infuse language models with additional embeddings for individual text segments. Concretely, we employ a self-attention-based pointer network to reconstruct the original order of a randomly shuffled sequence of text segments. This non-trivial pre-training task becomes exponentially more complex as the number of segments increases. Given a document of $N$ consecutive text segments the number of possible segment permutations grows factorially to $N!$. This inherent complexity requires the model to gain a deep contextual understanding, picking up on nuanced intricacies like coherence, chronological order, and causal relationships to ensure that the narrative flows logically and maintains continuity from beginning to end. 

To address the fact that transformer-based language models \cite{vaswani2017attention} typically exhibit an upper limit on the maximum token context size\footnote{The self-attention mechanism incurs a computational cost that scales quadratically in sequence length, imposing practical limits on the processable context size.}, denoted as $C$, we start with dissecting long text documents into individual training samples. Concretely, a training sample consists of $K$ text segments $s$, where $K$ is the maximum number of segments that fit within the language model's context window. Each segment is appended with a special delimiting token, [SEP], that indicates the segment's end. Note the value of $K$ varies for each sample, depending on the token length of the individual segments.

We enable the segment ordering task by randomly shuffling the segments within each training sample before encoding the entire sequence with a bidirectional language model denoted as \texttt{BiLM}. Specifically, we first apply WordPiece \cite{schuster2012japanese} tokenization to transform an examplary input sample consisting of $K$ segments into a sequence of sub-word tokens $t = (\text{[CLS]}, s_1, \text{[SEP]}_1, s_2, \text{[SEP]}_2, \dots,$ $s_K, \text{[SEP]}_K)$. [CLS] denotes the special start of sequence token and an individual segment $s_i = (t_1, \dots, t_m)$ consists of $m$ sub-word tokens, where $m$ can differ between segments. 

We couple our segment ordering pre-training task with masked language modeling (MLM) to enhance the model's understanding of context and word relationships. In line with the insights from \cite{cui2021pre} and \cite{wettig2023should}, we implement whole word masking and 
mask 15\% of randomly selected whole words. For the remaining MLM pre-training methodology of predicting the correct sub-words from a given vocabulary for all masked tokens we refer to \cite{Devlin19}.

The tokenized, masked, and permuted input sequence is then encoded by a \texttt{BiLM}, which yields a series of $d$-dimensional hidden state vectors $\bm{H} = (\bm{h}_1, \bm{h}_2, \ldots, \allowbreak \bm{h}_T) \in \RR^{T \times d}$ corresponding to each token $t_i$ for a sequence of length $T$.

For the segment reordering task, we collect the hidden states corresponding to the [SEP] tokens, denoted as $ \bm{H}_\text{[SEP]} = (\bm{h}^1_\text{[SEP]}, \bm{h}^2_\text{[SEP]}, \ldots, \bm{h}^K_\text{[SEP]}) \in \RR^{K \times d}$. We add learnable absolute positional embeddings $\bm{E} = (\bm{e}_1, \bm{e}_2, \ldots, \bm{e}_K)$ to each segment hidden state, yielding the enhanced segment representations $\bm{H}'_\text{[SEP]} = \bm{H}_\text{[SEP]} + \bm{E}$. Naturally, the added positional bias encodes the new segment position after shuffling, preventing the reordering task from being compromised.

Subsequently, we pass $\bm{H}'_\text{[SEP]}$ to a pointer network \cite{NIPS2015_29921001}, which is particularly suited for our SO task due to the varying number of segments per sample $K$, which precludes the use of a static output layer with a fixed number of classes. The network calculates each segment's probability distribution over the original segment positions using a multiplicative self-attention mechanism, defined as follows:
\begin{align}
\bm{A} = \text{softmax}\left(\frac{\bm{Q}\bm{K}^\top}{\sqrt{q}} \right), \quad \bm{Q} = \bm{H}'_\text{[SEP]} \bm{W}^\top_{\text{query}}, \quad \bm{K} = \bm{H}'_\text{[SEP]} \bm{W}^\top_{\text{key}},  \\ 
\bm{W}_{\text{query}} \in \RR^{q \times d}, \; \bm{W}_{\text{key}} \in \RR^{q \times d}, \; \bm{Q} \in \RR^{K \times q}, \; \bm{K} \in \RR^{K \times q}, \; \bm{A} \in \RR^{K \times K},
\end{align}
where $\bm{W}_{\text{query}}$ and $\bm{W}_{\text{key}}$ are learnable query and key weight matrices, $q = d / 4$ is their respective row dimension, and $\bm{A}$ is the row-stochastic\footnote{$\sum_{j=1}^{K} a_{ij} = 1 \quad \forall i \in \{1, 2, \ldots, K\}$.} square attention matrix, with each element $a_{ij}$ denoting the predicted probability that segment $i$ originated from position $j$. It follows that the predicted position of segment $i$ equals $\hat{y}_i = \argmax_{j}(\bm{a}_i)$.

The loss for the segment ordering task is computed using negative log-likelihood, $\mathcal{L}_{\text{SO}}(\bm{A}, \bm{y}) = -\sum_{i=1}^{K} \log(a_{i,y_i})$,
where $y_i$ denotes the ground truth position of segment $i$.

\subsection{Sample-efficient Fine-Tuning using Dynamic Sampling} \label{sec:dynamic-sampling}

Based on the previously detailed concept of combining multiple text segments to improve contextual understanding, this section focuses on the associated benefits in sample efficiency and introduces dynamic sampling to enhance data diversity.

Traditional text classification fine-tuning approaches for encoder-only language models like BERT often treat each text segment as an independent sample, which can lead to sub-optimal context utilization and unnecessary computational overhead, especially for short segments. Our method dynamically combines multiple text segments into a single sample, thereby maximizing the use of the model's context capacity $C$ and enhancing training efficiency.

For an average segment length of $\bar{T}$ tokens, the maximum number of segments per sample $K = \left\lfloor C / \bar{T} \right\rfloor$ represents the efficiency gain factor, which quantifies the improvement over processing segments individually. This gain is more evident when using large batch sizes $B$, as the longest sample in a batch dictates the memory and computational requirements. 

A drawback of uniting multiple segments in one sample is the reduction in sample diversity, which is particularly problematic in small datasets. To mitigate this issue and promote sample diversity, we introduce dynamic sampling for fine-tuning in scenarios with scarce data. Instead of deterministically merging the maximum number of consecutive segments $K$, we randomly select the number of combined segments $L$, sampling from a uniform distribution $\mathcal{U}(L_\text{min}, K)$, where $L_\text{min}$ denotes the minimum number of merged segments. While this reduces the expected computational efficiency gain per sample,
it introduces beneficial randomness into the training process. By exposing the model to varying consecutive segment combinations of different length during each epoch, we encourage better generalization and reduce the risk of overfitting.

Subsequently, our experiments validate that combining segment order pre-training with sample-efficient fine-tuning using dynamic sampling significantly enhances performance in downstream text classification tasks that require a comprehensive understanding of complex and extended document structures.

\section{Experiments} \label{sec:experiments}

We split our experiments in two parts. First, we focus on our pointer-guided pre-training setup before quantitatively evaluating its impact on five downstream tasks requiring sequential text classification. 

The experiments were conducted on a GPU cluster equipped with eight 32GB Nvidia Tesla V100 GPUs. The cumulative pre-training duration of all models amounted to 380 GPU hours. Our code is implemented in PyTorch, with the initial weights of pre-trained models being loaded from HuggingFace. We open-source our code base on \href{https://github.com/LarsHill/pointer-guided-pre-training}{GitHub}\footnote{\url{https://github.com/LarsHill/pointer-guided-pre-training}.}.

\subsection{Pre-Training}
In the following, we briefly introduce our pre-training datasets and discuss the overall training setup including baselines and results.

\subsubsection{Data}
Table \ref{tab:pretraining-data} details the mixture of our pre-training datasets and reports various descriptive statistics like the number of tokens and segments per dataset. 
\begin{table*}[t]
\caption{Descriptive statistics of pre-training datasets with document, segment, sample, and token counts in English and German, including total and average values (M\,= million, B = billion). Token and sample statistics are calculated based on the multi-lingual word-piece vocabulary, custom-100K (see Table \ref{tab:pretraining_results}), created from all pre-training datasets.}
    \centering
    \sisetup{
  group-separator={,}, 
  group-minimum-digits=4, 
  table-align-text-post=false, 
}
\setlength{\tabcolsep}{3.5pt}
\begin{tabular}{
  l 
  S[table-format=3.2] 
  S[table-format=3.2]
  S[table-format=4.2]
  S[table-format=3.2]
  S[table-format=3.2]
  S[table-format=3.2]
}
\toprule
 & \multicolumn{2}{c}{Wikipedia} & {Bundesanzeiger} & \multicolumn{2}{c}{News}  & {Sum} \\
 \cmidrule(lr){2-3}
 \cmidrule(lr){4-4}
 \cmidrule(lr){5-6}
  & {EN} & {DE} & {DE} & {EN} & {DE} & \\
\midrule
Documents (M) &  5.85  &  2.44 & 1.91 & 2.02 & 0.65 & 12.88\\
Segments (M) &  99.37 &  19.63& 85.06 & 36.25 & 17.51 & 257.81\\
Samples (M) & 13.83 & 4.80 & 9.53 & 5.23 & 1.23 & 34.62\\
Tokens (B) & 4.48 & 1.62 & 3.74 & 1.97 & 0.44 & 12.24\\
\midrule
Tokens (\%)  & 36.56 & 13.26 & 30.52 & 16.10 & 3.56 & 100\\
\bottomrule
\end{tabular}

    \label{tab:pretraining-data}
\end{table*}

The \texttt{Wikipedia} datasets comprise 5.9 (English) and 2.4 (German) million articles respectively that were directly retrieved from their rendered HTML pages. In contrast to the commonly used Wikimedia XML dumps, our corpus resolves Wikipedia's templating syntax embedded in the dumps, and thus represent the articles in their original form leading to improved data quality.

\texttt{Bundesanzeiger} contains 1.9 million German corporate annual reports 
from the \href{https://www.bundesanzeiger.de/}{Bundesanzeiger}, a platform where German companies are mandated to publish their legally required documents. Compared to Wikipedia the average document length is roughly three times higher resulting in around $2,000$ tokens per report.

Lastly, we include two proprietary datasets of English (2 million) and German (650 thousand) \texttt{news} articles that raise the total number of pre-training tokens to 12.24 billion.

For all datasets, we employ a 90-10 split for training and validation. We parse raw HTML articles using the \href{https://lxml.de/index.html}{lxml Python library}, distinguishing headlines, paragraphs, tables, enumerations and more. 

\subsubsection{Training Setup and Results}\label{sec:pretraining_setup}

We evaluate the efficacy of our pointer-guided segment ordering (SO) task by pre-training and fine-tuning three variants of the BERT language model: BERT, RoBERTa, and our proposed PointerBERT. The BERT model adheres to the original design by \cite{Devlin19}, employing self-supervised MLM and next sentence prediction (NSP). RoBERTa \cite{Liu2019RoBERTa} modifies the BERT pre-training scheme by omitting NSP and maximizing the use of context by concatenating multiple text segments. PointerBERT extends RoBERTa with the inclusion of our SO task, as detailed in Section \ref{sec:segment-ordering}. Note the application of SO is architecture agnostic and can be used to generally enhance the paragraph-level contextual comprehension of bidirectional encoder language models like RNNs \cite{chung2014empirical}, DeBERTa \cite{he2023debertav} and Electra \cite{Clark2020ELECTRA}.

Table \ref{tab:pretraining_results} presents the pre-training configurations for each model variant. All variants are based on \href{https://huggingface.co/google-bert/bert-base-cased}{Google's BERT$_\text{BASE}$} architecture as encoding backbone, differing only in their tokenizer and vocabulary construction.
\begin{table*}[t]
    \caption{Training configurations and validation accuracies for all language model variations and their pre-training tasks, masked language modeling (MLM), next sentence prediction (NSP) and segment ordering (SO). The scores represent averaged batch accuracies across the validation set.}
    \centering
\setlength{\tabcolsep}{3.5pt}
\begin{tabular}{llcllccc}
\toprule
 &  &   &  &  &  \multicolumn{3}{c}{Accuracy $\uparrow$ (\%)} \\
\cmidrule(lr){6-8}
Architecture &  Datasets & \makecell{Pre-\\trained} & \makecell{Train \\ steps} &  Tokenizer & MLM & NSP & SO \\
\midrule
BERT  & wiki-en  & \xmark & \num{1e+5}  &  bert-cased & $30.14$ & $73.85$ & $-$ \\
RoBERTa  & wiki-en  & \xmark & \num{1e+5}  &  bert-cased &  $44.77$ & $-$ & $-$ \\
PointerBERT & wiki-en  & \xmark & \num{1e+5}  &  bert-cased &  $43.10$ & $-$ & $34.90$ \\
\midrule
RoBERTa           & all &  \xmark & \num{2e+5}  &  custom-100K  &  $57.66$ & $-$ & $-$ \\
PointerBERT & all & \xmark &  \num{2e+5}  &  custom-100K &  $55.11$ & $-$ & $39.49$\\
\midrule
PointerSciBERT           & wiki-en  & \cmark & \num{1e+5}  &  scibert-uncased &   $58.27$ &   $-$ & $52.45$ \\
PointerBERT             & all-de  & \cmark & \num{1e+5}  &  bert-cased$_\text{dbmdz}$ &    $73.62$ &   $-$ & $57.35$ \\
\bottomrule
\end{tabular}
    \label{tab:pretraining_results}
\end{table*}

Concretely, we distinguish three scenarios. First, we train each model from scratch on the English Wikipedia corpus (wiki-en) using Google's \href{https://huggingface.co/google-bert/bert-base-cased}{bert-base-cased tokenizer}. The validation results demonstrate that PointerBERT correctly reorders shuffled segments in 35\% of cases. Although this may appear modest, it is significantly higher than the baseline random guess accuracy of approximately $1/5040 \approx \num{2e-4}$, given an average of 7 segments per sample (see Table \ref{tab:pretraining-data}) and 5040 possible permutations. Importantly, the inclusion of SO does not compromise MLM performance, with only a marginal decrease in validation accuracy. Lastly, we observe that combining MLM with NSP significantly diminishes MLM performance, likely due to the reduced sample efficiency and the underutilization of the model's context capacity.

Second, we train RoBERTa and PointerBERT on the combined multilingual datasets, denoted ``all'', using a newly developed 100K token word-piece vocabulary created from the pre-training data. This step aims to assess the impact of SO in a multilingual context. 

Third, we build upon the pre-trained checkpoints of Allen AI's SciBERT \cite{beltagy2019scibert} and the German BERT model released by the MDZ Digital Library team (dbmdz)\footnote{\url{https://huggingface.co/dbmdz/bert-base-german-cased}.} and continue pre-training, incorporating both MLM and SO. Here we aim to demonstrate the applicability of SO for already pre-trained language models and show that continuous pre-training with MLM and SO not only further increases the MLM performance but also manages to induce improved paragraph-level text understanding thanks to pointer-guided SO.

\begin{figure}[t]
  \centering
  \includegraphics[width=\linewidth]{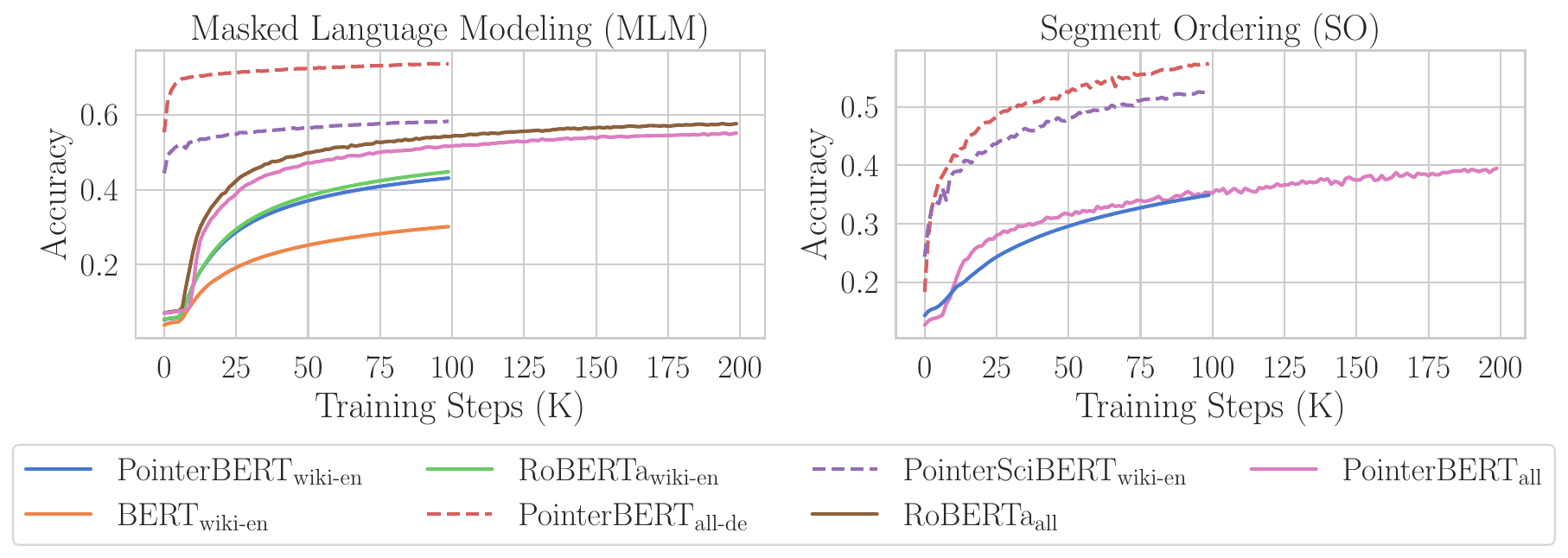}
  \caption{Pre-training progress for all model variants, showcasing validation accuracy curves for masked language modeling (MLM) and segment ordering (SO).}
  \label{fig:pretraining_results}
\end{figure}

All models are trained using gradient descent with the AdamW optimizer \cite{loshchilov2017decoupled}, featuring a 10\% linear warmup and a decaying learning rate schedule. We apply a weight decay of 0.01 and clip gradients at a maximum value of 1. The peak learning rate is set to \num{1e-4}, with a batch size of 16 and gradient accumulation over 4 steps, resulting in an effective batch size of 64. Model performance is evaluated on a hold-out validation set every $5,000$ steps. Figure \ref{fig:pretraining_results} provides a detailed view of the training progress, depicting both MLM and SO validation losses and accuracies.

Notably continued pointer-guided pre-training not only further improves MLM accuracy but also yields robust SO performance, which translates into enhanced results on downstream classification tasks, as discussed in the following section.

\subsection{Downstream Fine-Tuning for Sequential Text Classification}

In this section, we delve into the application of our pre-trained models to a series of fine-tuning downstream tasks that necessitate sequential text classification. We explore a diverse array of datasets, spanning both scientific literature and financial reporting domains, to assess the models' capabilities in categorizing text segments in different scenarios.

\subsubsection{Datasets}

Our evaluation encompasses five datasets, three from the scientific literature corpus and two from the financial reporting sector, each presenting unique challenges for text segment classification.

Within the scientific literature, we examine the \texttt{CSAbstruct} dataset \cite{cohan-etal-2019-pretrained}, which comprises 2,189 computer science abstracts with sentences annotated to discern their rhetorical roles. The \texttt{PubMed-RCT} dataset \cite{dernoncourt-lee-2017-pubmed} extends our evaluation to 20,000 biomedical abstracts from PubMed, segmented into five rhetorical categories, following the preprocessing methodology outlined by \cite{jin2018hierarchical}. The \texttt{Nicta} dataset \cite{kim2011automatic} further contributes with 1,000 biomedical abstracts, where sentences are classified according to the PICO framework (Population, Intervention, Comparison, Outcome) \cite{richardson1995well}.

Transitioning to the financial sector, we incorporate the \texttt{GRI$_\text{DE}$} dataset \cite{hillebrand_sustain}, which consists of 92 sustainability reports from leading German companies. These reports were initially obtained as PDFs from corporate websites and subsequently annotated by experts to correspond with the Global Reporting Initiative (GRI) standards, covering 89 indicators across economic, environmental, and social dimensions. The \texttt{IFRS$_\text{EN}$} dataset is composed of 45 English annual reports adhering to the International Financial Reporting Standards (IFRS). Provided by an auditing firm, these reports contain annotations that map paragraphs to 543 distinct legal requirements, with some paragraphs addressing multiple items.

A summary of the datasets' descriptive statistics is presented in Table \ref{tab:finetuning-data}.
\begin{table*}[t]
\caption{Descriptive statistics of scientific and financial fine-tuning datasets. Sample statistics are calculated based on the custom-100K 
vocabulary (see Table \ref{tab:pretraining_results}).
}
    \centering

\setlength{\tabcolsep}{3pt}
\begin{tabular}{lccccc}

\toprule
 &  \multicolumn{3}{c}{Scientific Abstracts} &  \multicolumn{2}{c}{Financial Reports} \\
\cmidrule(lr){2-4}
\cmidrule(lr){5-6}
 & CSAbstruct & PubMed-RCT & Nicta & IFRS$_{\text{EN}}$ & GRI$_{\text{DE}}$ \\
\midrule
Documents            & 2189 & 20,000 & 1000 & 45 & 92\\
Segments & 14,708 & 240,386 & 9771 & 19,573 & 89,412 \\
Samples & 2191 & 23,289 & 1061 & 4773 & 21,306 \\
\midrule
Is multi-label & \xmark & \xmark & \xmark & \cmark & \cmark \\
Classes & 5 & 5 & 6 & 543 & 89 \\
Segments labeled (\%) & 100.00 & 100.00 & 100.00 & 78.37 &  8.57 \\
\bottomrule
\end{tabular}
    \label{tab:finetuning-data}
\end{table*}
The scientific abstract datasets (CSAbstruct, PubMed-RCT, and Nicta) contrast with the financial report datasets (GRI$_\text{DE}$ and IFRS$_\text{EN}$) in terms of structure. The former category includes a higher volume of documents, each with approximately 10 sentences, typically fitting within the model's context window of 512 tokens. This is reflected in the average number of samples per document being close to 1 for these datasets. They feature a smaller set of categories and require multi-class classification, where every text segment is annotated and assigned to a single category.

Conversely, the financial datasets contain fewer but significantly longer documents, averaging over 400 segments. The resulting classification complexity is further amplified by a larger number of checklist categories and annotation scarcity, with only 8.5\% of paragraphs in the GRI$_\text{DE}$ dataset linked to a GRI requirement. Moreover, both datasets contain segments that refer to multiple checklist items, making this task a multi-label classification challenge that resembles the difficulty of information retrieval due to the severe class imbalance and annotation sparsity. 

\begin{figure}[t]
  \centering
  \includegraphics[width=\linewidth]{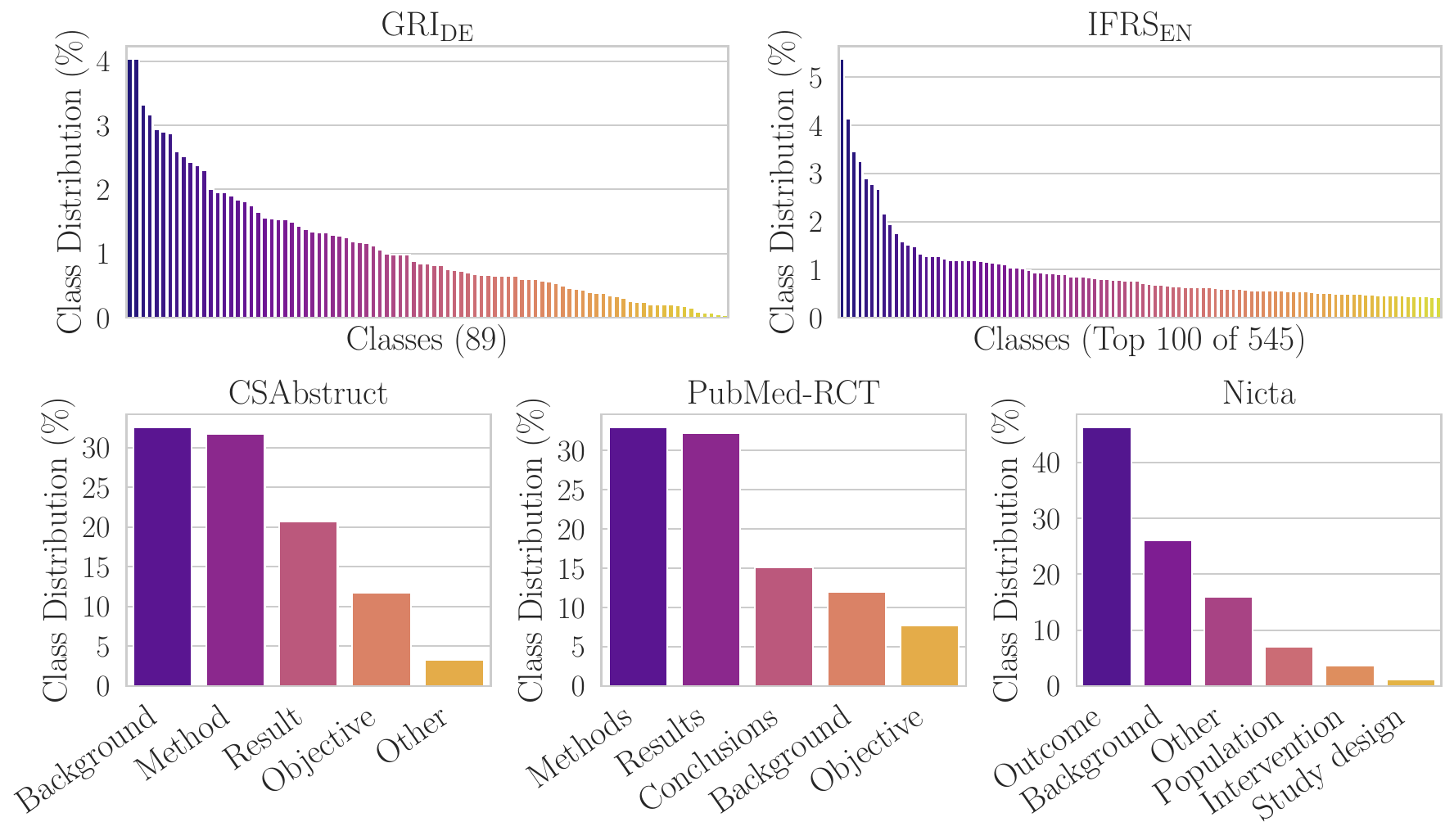}  
  \caption{Class distributions across all datasets showcasing label imbalances.}
  \label{fig:fine_tuning_class_distribution}
\end{figure}

The variation in class distribution across all datasets is graphically depicted in Figure \ref{fig:fine_tuning_class_distribution}.
For all datasets, we adhere to the established training, validation, and test splits introduced in prior work. For the new IFRS$_\text{EN}$ dataset, we employ a random split of 35 training, 5 validation, and 5 test documents.

\subsubsection{Baselines and Classification Tasks}

In the following comparative analysis, we benchmark our PointerBERT model variants against the various pre-trained baselines described in Section \ref{sec:pretraining_setup}. In contrast to the other baselines, BERT$_\text{wiki-en}$ processes each text segment individually and we utilize the hidden state vector of its special classifier token [CLS] as input for the downstream classification head. The remaining RoBERTa-based models handle multiple segments at a time, which is why we leverage the hidden states of their segment separating special tokens [SEP] for subsequent category prediction. 
The final output layer of each model differs depending on the dataset and its respective classification task. The scientific abstract datasets require multi-class classification which implies a softmax output layer that uniquely maps each segment $s_i$ to a single category $r_j \in \mathcal{R}$. Conversely, the financial report datasets necessitate a sigmoidal output layer for multi-label classification, where a segment $s_i$ receives relevance scores for all checklist requirements in $\mathcal{R}$.

We also compare our methodology with external state-of-the-art models that have not been pre-trained by us. For the scientific abstract datasets, we draw comparisons with \cite{cohan-etal-2019-pretrained} who utilize a pre-trained SciBERT \cite{beltagy2019scibert} model and fine-tune it in the fashion of our RoBERTa baselines using the model's [SEP] tokens as input for the classification layer. Additionally, we report results for the latest sequential sentence classification models \cite{brack2022crossdomain,jin2018hierarchical,shang2021span,yamada2020sequential} that are particularly optimized for incorporating sequential label dependencies while decoding. 
Lastly, we compare our methodology with a dedicated recommender model for sustainability reports that was introduced by \cite{hillebrand_sustain}. It 
leverages a pre-trained BERT architecture equipped with a multi-layer perceptron (MLP) to identify the most relevant checklist requirements for each text segment in the GRI$_\text{DE}$ dataset.

\subsubsection{Training Setup}

\begin{table}[t]
    \caption{Selected hyperparameters per dataset for our PointerBERT models based on the best validation-set micro F$_1$ and MAP@3 performances.}
    \centering

\setlength{\tabcolsep}{3pt}
\begin{tabular}{
    l
    c
    c
    c
    c
    c
}

\toprule
Hyperparameter      & CSAbstruct & PubMed-RCT & Nicta & IFRS$_{\text{EN}}$ & GRI$_{\text{DE}}$ \\
\midrule
Dropout             & 0.1        & 0.2 & 0.2 & 0.2 & 0.2 \\
Batch size          & 8          & 4 & 4 & 4 & 4 \\
Learning rate       & \num{1e-4} & \num{5e-5} & \num{5e-5} & \num{5e-5} & \num{1e-5} \\
Epochs              & 2          & 2 & 3 & 30 & 3 \\
Loss weighting      & \xmark     & \xmark & \xmark & \cmark & \cmark \\
Random oversampling & \xmark   & \xmark & \xmark & \xmark & \cmark \\
Dynamic sampling    & \xmark   & \xmark & \xmark & \cmark & \cmark \\
Classification head & RNN       & Linear & Linear & Linear & Linear \\
Label embedding dim & 32 & $-$ & $-$ & $-$ & $-$ \\

\bottomrule
\end{tabular}
    \label{tab:hyperparameters}
\end{table}
For all models and datasets, we conduct an exhaustive grid search across a wide range of hyperparameters to identify the best parameter combinations, evaluated on the hold-out validation set micro F$_1$ (scientific abstracts) and mean average precision (MAP) @3 scores (financial documents). Initially, we perform a broad parameter sweep to establish a viable starting point for each architecture and dataset, followed by a more detailed fine-tuning within the proximity of these initial parameters. The outcome of this rigorous process is detailed in Table \ref{tab:hyperparameters}, presenting the optimal configurations for our PointerBERT model across datasets.

In the following, we highlight a few insights from Table \ref{tab:hyperparameters}. First, we compare different classification heads for the single-label prediction tasks. Besides a standard linear output layer that classifies each segment simultaneously, we evaluate the performance of employing a gated recurrent unit (GRU) \cite{cho2014learning} that incorporates the previously predicted label information for the subsequent segment prediction (see 
\cite{hillebrand2022kpi} for more details).
Surprisingly, this more elaborate decoding method only slightly improves results for CSAbstruct, which indicates that label dependencies are already sufficiently encoded in the separator token hidden states.

Second, we mitigate the challenges of annotation scarcity and data imbalance in the financial report datasets, by utilizing class-based loss weighting, adjusting the binary cross entropy loss according to the segments' inverse class frequencies. Following \cite{hillebrand2022kpi}, we also adopt random oversampling for the GRI$_\text{DE}$ dataset, enhancing model exposure to annotated segments. 

Third, we employ dynamic segment sampling (Section \ref{sec:dynamic-sampling}), which increases sample diversity across epochs for models that are capable of processing multiple segments together. This sampling technique proves especially useful for the financial report datasets characterized by a small number of long documents.
For our experiments we set the minimum number of randomly selected segments per sample to $L_\text{min} = 3$. We refrain from applying dynamic sampling on the scientific abstract datasets because of their substantially larger size and almost all documents comfortably fitting within our models' 512-token context window.  

All additional training parameters not specified in Table \ref{tab:hyperparameters} align with the pre-training configurations described in Section \ref{sec:pretraining_setup}, except for gradient accumulation, which is not needed during fine-tuning due to smaller batch sizes. Also, to ensure a level playing field, all pre-trained baseline models undergo the same hyperparameter selection process and benefit from the described training enhancements, which enables fair test-set evaluations.

\subsubsection{Results} \label{sec:results}

Table \ref{tab:finetuning_results} presents the test-set performance metrics for all evaluated models across each dataset. 
\begin{table}[t]
    \caption{Test set results for sequential text classification on scientific abstract and financial document datasets. PointerBERT outperforms all competing baselines in micro and macro F$_1$ score as well as top 3/5 mean average precision (MAP). We report mean (best scores in bold) and standard deviation values from 10 independently seeded training runs for robust test set evaluation.}
    \centering
\begin{tabularx}{\textwidth}{Xcccccccccc}
\toprule
 &  \multicolumn{6}{c}{Scientific Abstracts} &  \multicolumn{4}{c}{Financial Reports} \\
\cmidrule(lr){2-7}
\cmidrule(lr){8-11}
 &  \multicolumn{2}{c}{CSAbstruct} &  \multicolumn{2}{c}{PubMed-RCT} &  \multicolumn{2}{c}{Nicta} &  \multicolumn{2}{c}{IFRS$_{\text{EN}}$} &  \multicolumn{2}{c}{GRI$_{\text{DE}}$}\\
\cmidrule(lr){2-3}
\cmidrule(lr){4-5}
\cmidrule(lr){6-7}
\cmidrule(lr){8-9}
\cmidrule(lr){10-11}
in \% &  \multicolumn{2}{c}{F$_1$} &  \multicolumn{2}{c}{F$_1$} &  \multicolumn{2}{c}{F$_1$} &  \multicolumn{2}{c}{MAP} &  \multicolumn{2}{c}{MAP} \\
\cmidrule(lr){2-3}
\cmidrule(lr){4-5}
\cmidrule(lr){6-7}
\cmidrule(lr){8-9}
\cmidrule(lr){10-11}
Architecture &  Micro & Macro & Micro & Macro & Micro & Macro &  @3 & @5 &  @3 & @5 \\
\midrule
Jin et al. \cite{jin2018hierarchical} & $81.30$ & $-$ & $92.60$ & $-$ & $84.70$ & $-$ & $-$ & $-$ & $-$ & $-$ \\
Cohan et al. \cite{cohan-etal-2019-pretrained} & $83.10$ & $-$ & $92.90$ & $-$ & $84.80$ & $-$ & $-$ & $-$ & $-$ & $-$ \\  
Yama. et al. \cite{yamada2020sequential} & $-$ & $-$ & $93.10$ & $-$ & $84.40$ & $-$ & $-$ & $-$ & $-$ & $-$ \\
Shang et al. \cite{shang2021span} & $-$ & $-$ & $92.80$ & $-$ & $\bm{86.80}$ & $-$ & $-$ & $-$ & $-$ & $-$ \\
Brack et al. \cite{brack2022crossdomain} & $-$ & $-$ & $93.00$ & $-$ & $86.00$ & $-$ & $-$ & $-$ & $-$ & $-$ \\
\makecell[lc]{Pointer- \\ \hspace{0.5em}SciBERT$_\text{wiki-en}$}             & \makecell[c]{$\bm{83.21}$ \\ \scriptsize{$\pm 0.31$}} & \makecell[c]{$\bm{82.05}$ \\ \scriptsize{$\pm 0.57$}} &   \makecell[c]{$\bm{93.56}$ \\ \scriptsize{$\pm 0.10$}} & \makecell[c]{$\bm{89.56}$ \\ \scriptsize{$\pm 0.16$}}  &   \makecell[c]{$85.07$ \\ \scriptsize{$\pm 0.33$}}  &   \makecell[c]{$76.22$ \\ \scriptsize{$\pm 1.13$}} &   \makecell[c]{$\bm{62.75}$ \\ \scriptsize{$\pm 1.00$}} & \makecell[c]{$\bm{63.99}$ \\ \scriptsize{$\pm 0.94$}} & $-$ & $-$ \\
\makecell[lc]{Hillebrand \\ \hspace{0.5em}et al. \cite{hillebrand_sustain}} & $-$ & $-$ &   $-$ & $-$  &   $-$  &   $-$ &   $-$ & $-$ & \makecell[c]{$33.37$ \\ \scriptsize{$\pm 0.95$}} & \makecell[c]{$35.29$ \\ \scriptsize{$\pm 0.91$}} \\
\makecell[lc]{Pointer- \\ \hspace{0.5em}BERT$_\text{all-de}$}            & $-$ & $-$ &   $-$ & $-$  &   $-$  &   $-$ &   $-$ & $-$ & \makecell[c]{$\bm{34.25}$ \\ \scriptsize{$\pm 1.07$}} & \makecell[c]{$\bm{36.19}$ \\ \scriptsize{$\pm 1.06$}} \\
\midrule
BERT$_\text{wiki-en}$               & \makecell[c]{$73.25$ \\ \scriptsize{$\pm 0.72$}} & \makecell[c]{$73.83$ \\ \scriptsize{$\pm 0.60$}} & 
\makecell[c]{$85.98$ \\ \scriptsize{$\pm 0.07$}} & \makecell[c]{$80.72$ \\ \scriptsize{$\pm 0.09$}} 
& \makecell[c]{$72.74$ \\ \scriptsize{$\pm 0.48$}} & \makecell[c]{$65.80$ \\ \scriptsize{$\pm 0.82$}} & \makecell[c]{$56.77$ \\ \scriptsize{$\pm 0.71$}} & \makecell[c]{$57.61$ \\ \scriptsize{$\pm 0.67$}} & $-$ & $-$ \\
RoBERTa$_\text{wiki-en}$            & \makecell[c]{$81.21$ \\ \scriptsize{$\pm 0.70$}} & \makecell[c]{$79.37$ \\ \scriptsize{$\pm 1.02$}} & \makecell[c]{$92.48$ \\ \scriptsize{$\pm 0.05$}} & \makecell[c]{$88.17$ \\ \scriptsize{$\pm 0.10$}} & \makecell[c]{$80.98$ \\ \scriptsize{$\pm 0.56$}} & \makecell[c]{$69.92$ \\ \scriptsize{$\pm 1.28$}} & \makecell[c]{$54.68$ \\ \scriptsize{$\pm 0.94$}} & \makecell[c]{$56.05$ \\ \scriptsize{$\pm 0.90$}} & $-$ & $-$ \\
\makecell[lc]{Pointer- \\ \hspace{0.5em}BERT$_\text{wiki-en}$} & \makecell[c]{$\bm{81.91}$ \\ \scriptsize{$\pm 0.41$}} & \makecell[c]{$\bm{80.42}$ \\ \scriptsize{$\pm 0.66$}} & \makecell[c]{$\bm{92.63}$ \\ \scriptsize{$\pm 0.06$}} & \makecell[c]{$\bm{88.29}$ \\ \scriptsize{$\pm 0.08$}} & \makecell[c]{$\bm{81.25}$ \\ \scriptsize{$\pm 0.27$}} & \makecell[c]{$\bm{70.82}$ \\ \scriptsize{$\pm 0.73$}} & \makecell[c]{$\bm{57.17}$ \\ \scriptsize{$\pm 1.17$}} & \makecell[c]{$\bm{58.43}$ \\ \scriptsize{$\pm 1.11$}} & $-$ & $-$ \\
\midrule
RoBERTa$_\text{all}$            & \makecell[c]{$81.60$ \\ \scriptsize{$\pm 0.46$}}   &  \makecell[c]{$79.92$ \\ \scriptsize{$\pm 0.72$}}  & \makecell[c]{$92.77$ \\ \scriptsize{$\pm 0.10$}}   &  \makecell[c]{$88.53$ \\ \scriptsize{$\pm 0.12$}}   &  \makecell[c]{$81.67$ \\ \scriptsize{$\pm 0.38$}} & \makecell[c]{$70.42$ \\ \scriptsize{$\pm 0.90$}} & \makecell[c]{$59.07$ \\ \scriptsize{$\pm 0.72$}} & \makecell[c]{$60.47$ \\ \scriptsize{$\pm 0.89$}} &  \makecell[c]{$27.83$ \\ \scriptsize{$\pm 1.79$}} & \makecell[c]{$29.88$ \\ \scriptsize{$\pm 1.91$}} \\
PointerBERT$_\text{all}$ &  \makecell[c]{$\bm{81.82}$ \\ \scriptsize{$\pm 0.71$}}  &   \makecell[c]{$\bm{80.36}$ \\ \scriptsize{$\pm 0.83$}} & \makecell[c]{$\bm{92.87}$ \\ \scriptsize{$\pm 0.10$}}  &   \makecell[c]{$\bm{88.65}$ \\ \scriptsize{$\pm 0.18$}} &    \makecell[c]{$\bm{81.92}$ \\ \scriptsize{$\pm 0.33$}} & \makecell[c]{$\bm{71.52}$ \\ \scriptsize{$\pm 1.06$}} & \makecell[c]{$\bm{59.50}$ \\ \scriptsize{$\pm 0.94$}} & \makecell[c]{$\bm{60.52}$ \\ \scriptsize{$\pm 0.87$}} & \makecell[c]{$\bm{28.84}$ \\ \scriptsize{$\pm 1.80$}} & \makecell[c]{$\bm{30.99}$ \\ \scriptsize{$\pm 1.75$}}\\
\bottomrule
\end{tabularx}

    \label{tab:finetuning_results}
\end{table}
To ensure reliability of our results, each model-dataset pairing has been evaluated 10 times using different seeds, with the average and standard deviation of these runs reported. We categorize the comparisons into three distinct groups. In the first category, our continuously pre-trained PointerBERT models are compared with current state-of-the-art models, revealing that the English PointerSciBERT model surpasses previous benchmarks on two out of three scientific abstract datasets. 
Notably, we did not focus on incorporating improved decoding mechanisms like advanced attention and CRF output layers, as included in the external baselines. Joining these methods with our PointerBERT methodology would likely further improve results. 
Additionally, the German PointerBERT$_\text{all-de}$ model exhibits enhanced retrieval performance on the German GRI$_\text{DE}$ sustainability report dataset. Due to the distinct vocabularies and monolingual pre-training approach, English models were not evaluated on German datasets and vice versa.

The second comparison group evaluates the PointerBERT architecture against RoBERTa and BERT models in a controlled setting. Both sets of models have been identically pre-trained from scratch, 
allowing any performance discrepancies to be attributed to architectural differences. This comparison underscores the superiority of our pointer-guided SO task, with PointerBERT consistently outperforming the other architectures across all English datasets.

In the final comparison group, the multilingual PointerBERT$_\text{all}$ model is pitted against the RoBERTa$_\text{all}$ baseline. Both models have been pre-trained from scratch as well using the same datasets, vocabulary, and training configurations (see Table \ref{tab:pretraining_results}. Evaluated across all five datasets, the PointerBERT$_\text{all}$ model consistently outperforms the RoBERTa$_\text{all}$ baseline, showcasing the efficacy of our pointer-guided architecture.

Overall, our findings demonstrate that the pointer-guided SO methodology, combined with dynamic sampling for efficient fine-tuning, surpasses all competing baselines across a diverse array of datasets and classification tasks.

\subsection{Limitations}

Despite the promising results, our current experiments have a few practical limitations that we plan to improve upon in future work. Firstly, the models are constrained by a context window size of only 512 tokens and employ absolute positional embeddings. Incorporating advanced attention mechanisms \cite{dao2022flashattention}, along with relative positional embeddings \cite{su2024roformer},
enables our pre-training approach to accommodate longer input sequences during inference. This enhancement will not only increase the complexity and effectiveness of the SO pre-training task but also enable the model to capture more distant paragraph-level context.

Besides scaling up our pre-training methodology in terms of larger model sizes, increased number of training steps and larger datasets, we also aim to extend our evaluation and fine-tuning efforts to information retrieval and semantic search tasks \cite{khattab2020colbert,reimers2019sentence}. 
Specifically, we seek to assess our methodology's effectiveness in identifying semantically relevant passages from long documents in response to natural language queries. We assume that our method's enhanced understanding of paragraph-level context and its ability to jointly embed subsequent text segments has the potential to improve semantic search and thereby retrieval augmented generation (RAG) \cite{lewis2020retrieval}.

\section{Conclusion}
\label{sec:conclusion}

We introduce a novel approach to enhance the contextual sensitivity of paragraph-level text representations through a pointer-guided segment ordering (SO) pre-training strategy and dynamic sampling for fine-tuning. Our methodology aims at improving the understanding of document structure and coherence, which is crucial for a wide range of downstream NLP applications, including text classification and information retrieval.

Our pre-training methodology leverages a self-attention-driven pointer network to restore the original sequence of shuffled text segments, thereby requiring the model to develop a deep understanding of narrative flow, coherence, and contextual relationships. This task, combined with masked language modeling, shows to significantly enhance the model's ability to comprehend and represent paragraph-level context. We further establish dynamic sampling during the fine-tuning phase to increase the diversity of training instances across epochs and improve sample efficiency. This sampling technique proves particularly beneficial for small datasets with long documents, as it helps to mitigate overfitting and to foster better generalization.

Our experiments demonstrate that models pre-trained with our pointer-guided SO task outperform existing baselines across a variety of datasets and tasks. Notably, our PointerBERT models achieve superior performance on both scientific literature and financial reporting datasets, showcasing the versatility and effectiveness of our approach. Looking ahead, we aim to overcome current limitations by incorporating more sophisticated language model backbones and broadening our evaluation framework to include information retrieval and semantic search tasks.

In conclusion, our work contributes an advancement in representation learning for paragraph-level text, setting a new benchmark for sequential text classification and paving the way for future research in document structure-based language modeling.

\begin{credits}
\subsubsection{\ackname} This research has been funded by the Federal Ministry of Education and Research of Germany and the state of North-Rhine Westphalia as part of the Lamarr-Institute for Machine Learning and Artificial Intelligence, LAMARR22B.

\subsubsection{\discintname}
The authors have no competing interests to declare that are relevant to the content of this article.
\end{credits}
%
%
%
\bibliographystyle{splncs04}
\bibliography{bibliography}
%




\end{document}